\documentclass[10pt,twocolumn,letterpaper]{article}

\usepackage{btas}
\usepackage{times}
\usepackage{epsfig}
\usepackage{graphicx}
\usepackage{amsmath}
\usepackage{amssymb}
\usepackage{booktabs, multicol, multirow}
\usepackage{array}
\usepackage{color}
\usepackage{subfigure}
\usepackage{caption}
\usepackage{cite}
\usepackage{balance}
\usepackage{comment}
\usepackage[norule,symbol,perpage]{footmisc}

\btasfinalcopy 
\newcommand*\samethanks[1][\value{footnote}]{\footnotemark[#1]}

\ifbtasfinal\fi
\makeatletter  
\def\ps@IEEEtitlepagestyle{  
\def\@oddfoot{\mycopyrightnotice}  
\def\@evenfoot{}  
}  
\def\mycopyrightnotice{  
{\hfill \footnotesize 978-1-7281-1522-1/19/\$31.00 \copyright 2019 IEEE\hfill}  
}  
\makeatother
\begin{document}

\title{Attribute-Guided Coupled GAN for Cross-Resolution Face Recognition}

\author{Veeru Talreja\thanks{Authors Contributed Equally}, \ Fariborz Taherkhani\samethanks[1], \ Matthew C. Valenti, and \ Nasser M. Nasrabadi\\
West Virginia University\\
Morgantown, WV, USA\\
{\tt\small vtalreja@mix.wvu.edu,fariborztaherkhani@gmail.com,valenti@ieee.org,nasser.nasrabadi@mail.wvu.edu }
}

\maketitle
\thispagestyle{empty}

\begin{abstract}
    In this paper, we propose a novel attribute-guided cross-resolution (low-resolution to high-resolution) face recognition framework that leverages a coupled generative adversarial network (GAN) structure with adversarial training to find the hidden relationship between the low-resolution and high-resolution images in a latent common embedding subspace. The coupled GAN framework consists of two sub-networks, one dedicated to the low-resolution domain and the other dedicated to the high-resolution domain. Each sub-network aims to find a projection that maximizes the pair-wise correlation between the two feature domains in a common embedding subspace. In addition to projecting the images into a common subspace, the coupled network also predicts facial attributes to improve the cross-resolution face recognition. Specifically, our proposed coupled framework exploits facial attributes to further maximize the pair-wise correlation by implicitly matching facial attributes of the low and high-resolution images during the training, which leads to a more discriminative embedding subspace resulting in performance enhancement for cross-resolution face recognition.  The efficacy of our approach compared with the state-of-the-art is demonstrated using the LFWA, Celeb-A, SCFace and UCCS datasets.  
\end{abstract}
\let\thefootnote\relax\footnotetext{\mycopyrightnotice}
\section{Introduction}

Facial biometrics is used in a variety of modern recognition and surveillance applications ranging from stand-alone camera applications in banks and supermarkets to multiple networked closed-circuit televisions in law enforcement applications or even in cloud-based authentication applications. \cite{talreja_icc_2019,talreja_globalsip_2017,talreja_2018_using, taherkhani_2018_facial, taherkhani_2018_deep,vtalreja_2018}. The large distance between surveillance cameras and the subjects leads to low-resolution (LR) face regions in the captured images. Usually, the discriminant properties of the face are degraded in the LR images, which leads to a significant drop in the accuracy of traditional face recognition algorithms developed for high-resolution (HR) images. An efficient face-recognition algorithm should perform well even for LR faces without significantly reducing recognition accuracy.   

In comparison to HR face images, LR faces have their own unique visual properties. Although many visual features are missing in LR face images, humans are still able to notice similarities between the LR and HR face images of a given subject. This implies that the neural systems of the human brain is able to  recover missing visual properties of LR faces if the human brain is familiar with
the high-resolution image of that subject or a given identity \cite{SKD}. Inspired by this fact, several LR face recognition models have been introduced that can be generally  divided into two categories: the hallucination category and the embedding category.
The models in the hallucination category reconstruct HR faces from LR faces before recognition \cite{kolouri2015transport,jian2015simultaneous,yang2015recognition,SHSR,uiboupin2016facial,10.1007/978-3-319-10593-2_13,DBLP:journals/corr/LedigTHCATTWS16}. The hallucination category of super-resolution is also used for other applications \cite{ferdous_2019_super}. For instance, Kolouri and Rohde \cite{kolouri2015transport} introduced a method based on optimal transport for single frame super-resolution to automatically build a nonlinear Lagrangian model of HR facial appearance. Thereafter, the LR facial image is improved by exploring the parameters of the model which perfectly fit the given LR data. 


Methods based on hallucination usually achieve  promising results in recognizing the reconstructed HR face images. However, the super-resolution operation in hallucination models usually requires significant additional computation that often translates to a reduction in the recognition speed. In contrast to methods based on the hallucination, methods in the embedding category extract  features from LR faces by leveraging various external face contexts.  Ren et al. \cite{ren2012coupled} introduce a coupled kernel embedding to implicitly map face images with different resolutions into an infinite  space. The recognition task is then performed in this new space to minimize the dissimilarities obtained by their kernel Gram matrices in the low and high-resolution spaces, respectively. Intuitively, the main step in the embedding method is to transfer knowledge from HR to LR face images. However, in these methods, one must be careful to transfer only the desired knowledge instead of  transferring all knowledge from a HR domain to a LR domain.

In addition to knowledge sharing between the high and low-resolution images, soft biometric traits such as facial attributes can also be used as complimentary information to improve the cross-resolution face recognition model. Facial attributes have been previously used jointly with face biometrics in different face recognition applications
\cite{survey_2016_dantcheva}. 



 In this paper, we present an embedding model for cross-resolution face recognition based on novel attribute-guided deep coupled learning framework using generative adversarial network (GAN) to find the hidden relationship between the features of high-resolution and low-resolution images in a latent common embedding subspace.  The framework also utilizes convolutional neural network (CNN) weight sharing followed by dedicated weights for learning the representative features for each specific face attribute. Specifically, our coupled framework exploits the facial attribute to further maximize the correlation between the low-resolution and high-resolution domains, which leads to a more discriminative embedding subspace to enhance the performance of the main task, which is the cross-resolution face recognition. Additionally, in our approach, we also predict the attributes for low-resolution images along with cross-resolution face recognition in a multi-tasking paradigm. Multi-task learning attempts to solve correlated tasks simultaneously by leveraging the knowledge sharing between the two tasks \cite{taherkhani_iet,taherkhani_aaai_2019,sobhan_btas}.  To summarize, our main contributions are:
\begin{itemize}
\vspace{-0.25cm}
\item A novel attribute guided cross-resolution (low-resolution to high-resolution) face recognition model using coupled GAN and multiple loss functions.
\vspace{-0.25cm}
\item A mutli-task learning framework to predict facial attributes for low-resolution facial images. 
\vspace{-0.25cm}
\item Extensive experiments using four different datasets and a comparison of the proposed method with state-of-the-art methods.  
    \end{itemize}

\section{Related Work}

\begin{figure*}[h]
\vspace{-0.30cm}
\centering
\includegraphics[width=13cm,height=10cm]{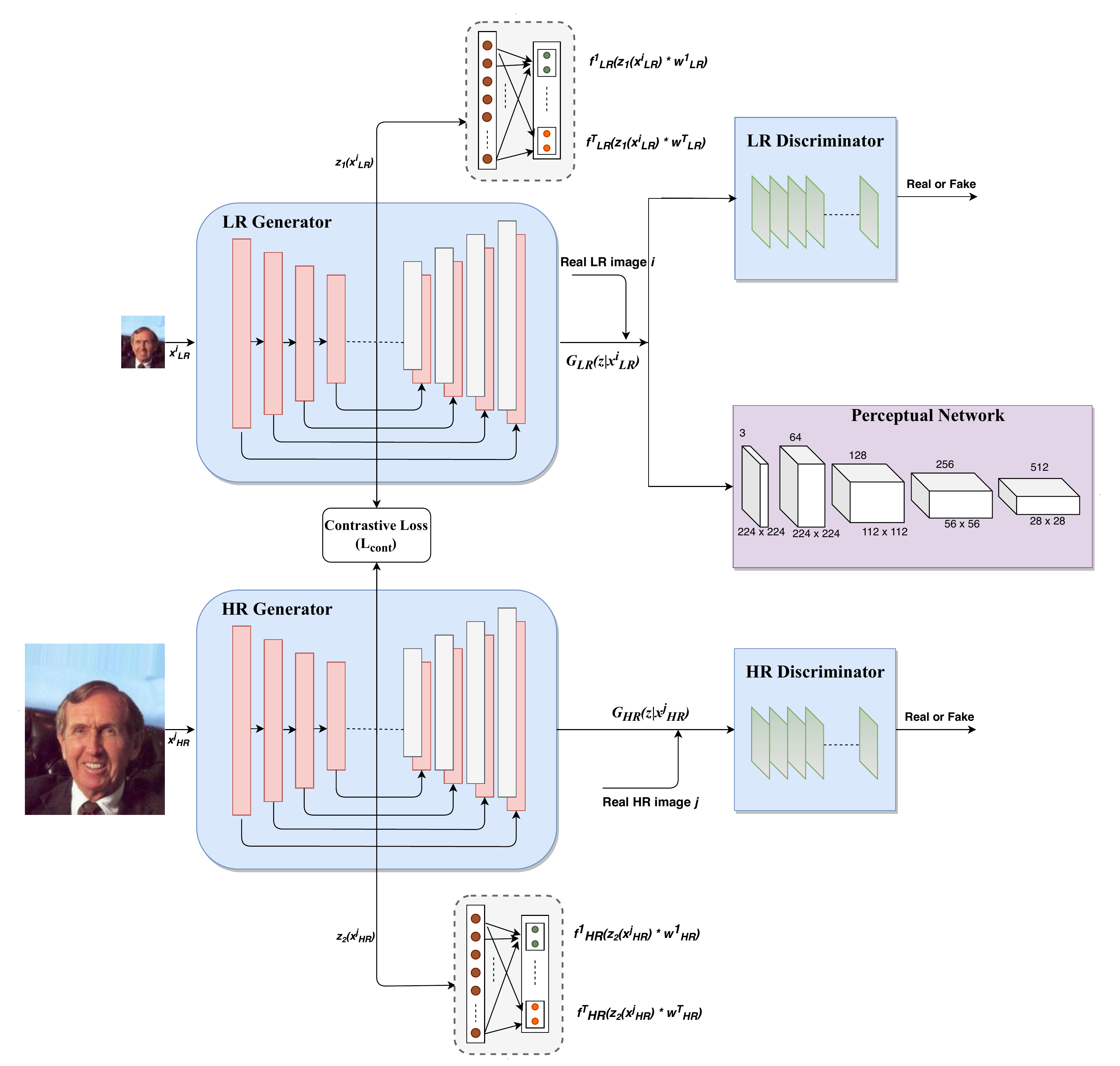}
\vspace{-0.20cm}
\caption{Block diagram of the proposed framework. 
}\label{fig:enrol}
\vspace{-0.35cm}
\end{figure*}

There are two categories of methods for low-resolution face recognition. The hallucination based methods \cite{uiboupin2016facial,10.1007/978-3-319-10593-2_13,DBLP:journals/corr/LedigTHCATTWS16,SHSR,yang2015recognition} reconstruct high-resolution faces before the face recognition, while embedding based methods extract latent features directly from low-resolution and high-resolution faces by using the embedding techniques. Yang et al. \cite{yang2015recognition} leverage sparse representation to simultaneously perform recognition and hallucination to synthesize person-specific versions of low-resolution faces without a significant drop in recognition. In \cite{uiboupin2016facial}, an algorithm is 
proposed to recognize faces via sparse representation
with a specific dictionary that includes many natural and facial images. Furthermore, deep models such as the ones presented in \cite{10.1007/978-3-319-10593-2_13} and \cite{DBLP:journals/corr/LedigTHCATTWS16} can generate
intensely realistic high-resolution images from low-resolution faces. Nevertheless, the speed of such super-resolution or hallucination methods might be a bit slow because of the complex high-resolution face reconstruction process, which restricts their direct use in applications where computational resources are limited. 

Rather than reconstructing high-resolution faces, a more
direct method is embedding low-resolution faces into different
external contexts to retrieve the missing information during resolution deterioration \cite{SKD}. Inspired by this, embedding methods have been proposed to transform both the high and low-resolution faces into a integrated feature domain for matching \cite{shekhar2017synthesis,zhang2015coupled,jiang2016cdmma,mudunuri2016low,wang2016pose,xing2016couple,wang2016studying,haghighat2017low,Li_2019_LRFRW,D-align_2019_TIFS,SKD}. In \cite{mudunuri2016low}, the multi-dimensional scaling is used to learn a common transformation matrix to jointly transform the facial features of low and high-resolution training face images. On the other hand, Wang et al. \cite{wang2016studying} solve very low-resolution recognition problem via deep learning approaches. In \cite{herrmann2016low}, CNNs are used with a manifold-based track comparison technique for low-resolution face recognition in videos. It is worth mentioning that the core idea of the embedding-based models is to transfer the knowledge from high-resolution face images, which is also the main idea of our proposed method.  


\section {Generative Adversarial Network}

Generative adversarial networks (GANs) have been widely used in different computer vision application such as style transfer, sketch to photo synthesis, and also in military applications \cite{sobhan_icb, osahor_2019_design,kazemi_nips_2018,kazemi_wacv_2019,kazemi_biosig_2018,osahor_2019_deep}. GANs  consist of two competing networks, namely a generator G and discriminator D. The goal of GAN is to train the generator G to produce samples from training noise distribution $p_{z}(z)$ such that the discriminator D cannot distinguish the synthesized samples from actual data $y$ with distribution $p_{data}$. Generator $G(z;\theta_{g})$ is a differentiable function which maps the noise variable $z$ to a data space using the parameters $\theta_{g}$. On the other hand, discriminator $D(.;\theta_{d})$ is also a differentiable function, which tries to discriminate using a binary classification between the real data $y$ and $G(z)$. Specifically, the generator and discriminator compete with each other in a two-player minimax game to minimize the Jenson-Shannon divergence \cite{NIPS_GAN}. The loss function $L(D,G)$ for GAN is given as:
 \vspace{-0.30cm}
\begin{equation}\begin{split}
     L(D,G) & = E_{y\sim P_{data}(y)}[\log D(y)]\\ & + E_{z\sim P_{z}(z)}[\log (1-D(G(z)))] \end{split}
 \end{equation}
 The objective (two player minimax game) for GAN is given by:
  \vspace{-0.30cm}
 \begin{equation}\begin{split}
     \min_{G}\max_{D} L(D,G) & =\min_{G}\max_{D}[E_{y\sim P_{data}(y)}[\log D(y)]\\ & + E_{z\sim P_{z}(z)}[\log (1-D(G(z)))]]\end{split}\label{eq:2}
 \end{equation}
 
 Conditional GAN is another variant of GAN where both the generator and the discriminator are conditioned on an additional variable $x$. This additional variable could be any kind of auxilary information such as discrete labels \cite{mirza_2014_conditional}, text \cite{RAYLLS_16}, or images \cite{Isola_2017_ImagetoImageTW}. The loss function for conditional GAN is given as:
 
 \begin{equation}\begin{split}
     L_{c}(D,G) & = E_{y\sim P_{data}(y)}[\log D(y|x)]\\ & + E_{z\sim P_{z}(z)}[\log (1-D(G(z|x)))].\end{split}\label{eq:3}
 \end{equation} The objective for the conditional GAN is the same two player minimax game as in (\ref{eq:2}) with loss function as $L_{c}(D,G)$ . Hereafter, we will denote the objective for conditional GAN as $O_{cGAN}(D,G,y,x)$, which is given by: \begin{equation}\begin{split}
O_{cGAN}(D,G,y,x) & = \min_{G}\max_{D} [E_{y\sim P_{data}(y)}[\log D(y|x)]\\ & + E_{z\sim P_{z}(z)}[\log (1-D(G(z|x)))]].\end{split}\label{eq:4}
 \end{equation}
 \vspace{-0.25cm}
\section{Proposed Method}
 In this section, we describe the proposed method for cross-resolution face recognition. In contrast to the hallucination approach, we do not up-sample each low-resolution image to the high-resolution domain before matching. Instead, we seek to project the high and low-resolution images to a common latent low-dimensional embedding subspace using generative modeling. Inspired by the success of GANs \cite{NIPS_GAN}, we explore adversarial networks in a multi-tasking paradigm to project low and high-resolution images to a common subspace for recognition, and also predict facial attributes from low recognition images.  
 
 As shown in Fig. \ref{fig:enrol}, the proposed method consists of a coupled framework made of two sub-networks, where each sub-network is a GAN architecture made of a generator and a discriminator. The generators are coupled together using a contrastive loss function. Each generator is also responsible to predict attributes in a multi-tasking paradigm. In addition to the adversarial loss, and contrastive loss, we propose to guide the sub-networks using a perceptual loss based on the VGG 16 architecture and also an $L_2$ reconstruction error. This is because  the perceptual loss in optimization helps to achieve a realistic image reconstruction  \cite{Johnson2016PerceptualLF}. 
 
 
 \subsection{Deep Coupled Framework}
 
 The objective of our method is the recognition of low-resolution face images with respect to a gallery of high-resolution images, which have not been seen during the training. The matching of the low-resolution and the high-resolution images is performed in a common embedding subspace. For this reason, we use a coupled framework which contains two sub-networks: low-resolution (LR) network and high-resolution (HR) network. The LR network consists of a GAN (generator + discriminator), attribute predictor, and a perceptual network based on VGG-16, while the HR network consists of a GAN (generator + discriminator) and an attribute predictor. 
 
 For the generators, we have used a U-Net network \cite{ronneberger_2015_unet} to better capture the low-level features and overcome the vanishing gradient problem due to deep network.  Motivated by \cite{Isola_2017_ImagetoImageTW}, we have used patch-based discriminators, which are trained iteratively along with the respective generators. Patch-based discriminator ensures preserving of high-frequency details which are usually lost when only $L_1$ loss is used. The final objective of our proposed method is to find the global deep latent features in a common embedding subspace representing the relationship between the low-resolution and their corresponding high-resolution face images. To find this common subspace between the two domains, we couple the two generators via a contrastive loss function $L_{cont}$ \cite{Chopra2005LearningAS}. 
 
 This loss function  $(L_{cont})$ is minimized so as to drive the genuine pairs (i.e., a LR image with its own corresponding HR image) towards each other in a common embedding subspace, and at the same time, push the impostor pairs (i.e., a LR image of a subject with another subject's HR image) away from each other. Let $x^i_{LR}$ denote the input LR face image, and $x^j_{HR}$ denote the input HR image. $c(i,j)$ is a binary label, which is equal to 0 if $x^i_{LR}$ and $x^j_{HR}$ belong to the same class (i.e., genuine pair), and equal to 1 if $x^i_{LR}$ and $x^j_{HR}$ belong to the different class (i.e., impostor pair). Let $z_1(.)$ and $z_2(.)$ denote the deep convolutional neural network (CNN)-based embedding functions to transform  $x^i_{LR}$ and $x^j_{HR}$, respectively into a common latent embedding subspace. Then, contrastive loss function $(L_{cont})$  if $c(i,j)=0$ (i.e., genuine pair) is given as:
\begin{equation}
\begin{split}
L_{cont}(z_1(x^i_{LR}),z_2(x^j_{HR}),c(i&,j)) = \\ & 
  \frac{1}{2}\left\lVert z_1(x^i_{LR})-z_2(x^j_{HR})\right\rVert^2_2.  
  \end{split}
  \end{equation}
Similarly if $c(i,j)=1$ (i.e., impostor pair), then contrastive loss function $(L_{cont})$ is :
  \begin{equation}
  \begin{split}
L_{cont}(z_1(x^i_{LR}),&z_2(x^j_{HR}),c(i,j))  = \\ & \frac{1}{2}\mbox{max}\biggl(0,m-\left\lVert z_1(x^i_{LR})-z_2(x^j_{HR})\right\rVert^2_2\biggr), 
\end{split}
\end{equation}
where $m$ is the contrastive margin and is used to ``tighten" the constraint. Therefore, the total loss function for coupling the sub-networks is denoted by $L_{cpl}$ and is given as: 



\vspace{-0.25cm}
\begin{equation}
    \begin{split}
        L_{cpl}=\frac{1}{N^2}\sum_{i=1}^{N}\sum_{j=1}^{N}L_{cont}(z_1(x^i_{LR}),z_2(x^j_{HR}),c(i,j)), 
    \end{split}\label{eq:6}
\end{equation}
where N is the number of training samples. The main motivation for using the coupling loss is that it has the capacity to find the discriminative embedding subspace because it uses the class labels implicitly, which may not be the case with some other metric such as Euclidean distance. This discriminative embedding subspace would be useful for matching of the LR images with the HR images and also for attribute prediction task.

\subsection{Attribute Prediction Task}

In addition to cross-resolution face recognition, another important objective of our proposed method is prediction of attributes using a LR or HR face image. However, separating these two objectives by learning multiple CNNs individually is not optimal since different objectives may share common features and have hidden relationship, which can be leveraged to jointly optimize the objectives. This notion of joint optimization has been used in \cite{Zhong2016FaceAP}, where they train a CNN for face recognition, and utilize the features for attribute prediction. Therefore, for this task, we use the respective feature set (i.e., $z_1(x^i_{LR})$ for LR, or $z_1(x^j_{HR})$ for HR)  from the common embedding subspace to also predict the attributes for a given image. Also, our network shares a large portion of its parameters among different attribute prediction tasks in order to enhance the performance of the recognition task in a mutli-task paradigm.  

 For the attribute prediction task, a LR or HR image is given as input to the network to predict a set of attributes. Consider that the input is a LR image denoted by $x^i_{LR}$, where the class label for the image is given by $\ell^{i} \in L$ for $i=1,\cdots,N$ where $N$ is the number of training samples. Let's consider $T$ to be the number of different facial attributes and $a^{i,t}$ denotes the ground truth attribute label for training sample $i$ and attribute $t$ for $t=1 \cdots T$. In this case, using the feature set from the common embedding subspace, the attribute prediction loss function is given as:
 \begin{equation}
     \begin{split}
         L_{aLR}=\frac{1}{N}\sum_{i=1}^{N}\sum_{t=1}^{T}l(f^t_{LR}(z_1(x^i_{LR})\times w^t_{LR}),a^{i,t}),
     \end{split}\label{eq:7}
 \end{equation}
where $f^t_{LR}(.)$ is a binary classifier for the attribute $t$ operated on the bottle-neck of LR generator as shown in Fig. \ref{fig:enrol}. The classifier is learned by using a loss function $l$ (e.g., cross entropy) and $w^t_{LR}$ represents the weight parameters for the classifier and these parameters are learned separately for each facial attribute task. 

Similarly, we can consider the other sub-network (HR network) and perform the same procedure with a HR image. The HR network also predicts the set of facial of attributes using the features $z_2(x^j_{HR})$  from the common embedding subspace for a given HR image. Therefore, the loss function for facial attribute prediction for HR image is given as:
\vspace{-0.25cm}
\begin{equation}
     \begin{split}
         L_{aHR}=\frac{1}{N}\sum_{j=1}^{N}\sum_{t=1}^{T}l(f^t_{HR}(z_2(x^j_{HR})\times w^t_{HR}),a^{j,t}),
     \end{split}
 \end{equation}
 where the notations are similar to (\ref{eq:7}) but correspond to the HR network. The total attribute prediction loss given as:
 \begin{equation}
     \begin{split}
         L_{a}=L_{aLR}+L_{aHR}.
     \end{split}
 \end{equation}

\subsection{Generative Adversarial Loss}

Let $G_{LR}$ and $G_{HR}$ denote the generators that synthesize the corresponding LR and HR images from the input LR  and HR image, respectively. Let $D_{LR}$ and $D_{HR}$ denote the discriminators for LR and HR GANs respectively. We have utilized the GAN loss function \cite{NIPS_GAN} to train the generators and the corresponding discriminators in order to ensure that the discriminators cannot distinguish the synthesized images by the generators from the corresponding ground truth images. Also, it can be observed from Fig. \ref{fig:enrol}, that the generators $G_{LR}$ and $G_{HR}$ try to generate a LR and HR image with the network conditioned on the input LR and HR image, respectively. The total loss for the coupled GAN is given by:

\vspace{-0.25cm}
\begin{equation}
L_{GAN}=L_{LR}+L_{HR},
\end{equation}
where $L_{LR}$ and $L_{HR}$ denote the GAN loss functions for the LR and the HR network, respectively and are given as:
\vspace{-0.25cm}
\begin{equation}
L_{LR}=O_{cGAN}(D_{LR},G_{LR},y^i,x^i_{LR})
\end{equation}
\begin{equation}
L_{HR}=O_{cGAN}(D_{HR},G_{HR},y^j,x^j_{HR}), 
\end{equation}where function $O_{cGAN}$ is given by (\ref{eq:4}). $x^i_{LR}$ ($x^j_{HR}$) is the LR (HR) image used as a condition for the LR (HR) GAN and $y^i$ ($y^j$) denotes the real LR (HR) data.  Note that the real LR (HR) data $y^i$ ($y^j$) and the network condition given by $x^i_{LR}$ ($x^j_{HR}$) are the same.




\subsection{Perceptual Loss}\label{subsec:percloss}
Perceptual loss function was introduced in \cite{Johnson2016PerceptualLF} for style transfer and super-resolution. In \cite{Johnson2016PerceptualLF}, instead
of relying only on $L_1$ or $L_2$ reconstruction error, the network parameters are learned using errors between high-level image
feature representations extracted from a pre-trained convolutional neural network. Similarly, in our proposed approach, perceptual loss is added only to the LR network using a pre-trained VGG-16 \cite{simonyan2014very} network to extract high-level features (ReLU3-3 layer) and the $L_1$ distance between these features of real and synthesized images is used to guide the generator $G_{LR}$. The perceptual loss for features for only the LR network is:
\vspace{-0.25cm}
\begin{equation}
    \begin{split}
        L_{P_{LR}}=&\frac{1}{C_pW_pH_p}\sum_{c=1}^{C_{p}}\sum_{w=1}^{W_{p}}\sum_{h=1}^{H_{p}} \\ & \left\lVert V(G_{LR}(z|x^i_{LR}))^{c,w,h}-V(y^i)^{c,w,h}\right\rVert,
    \end{split}
\end{equation}
where $y^i$ is the ground truth LR image, $G_{LR}(z|x^i_{LR})$ is the output of the LR generator. $V(.)$ represents a particular layer of the VGG-16 network, where the layer dimensions are given by $C_p$, $W_p$, and $H_p$. We have applied the perceptual loss only for the LR network to generate more sharper LR images helpful for recognition.

Similarly, we utilized perceptual loss for attribute prediction as well to measure the difference between the facial attributes of the synthesized and the real image. We applied the perceptual loss for attributes for both the LR and the HR network. To extract the attributes from a given HR image, we fine-tune the pre-trained VGG-Face \cite{parkhi2015deep} on 12 annotated facial attributes, which are shown in Table \ref{table:attr_pred}. After this, we utilize this attribute predictor to measure the attribute perceptual loss for both the LR and HR networks and the respective losses are given below:

\begin{equation}
    L_{pa_{LR}}= \left\lVert A(G_{LR}(z|x^i_{LR}))-A(y^i)\right\rVert_2^2,
\end{equation}
\begin{equation}
    L_{pa_{HR}}= \left\lVert A(G_{HR}(z|x^j_{HR}))-A(y^j)\right\rVert_2^2,
\end{equation}
where A(.) is the fine-tuned VGG-Face attribute predictor network. The total attribute perceptual loss is the sum of the perceptual attribute loss for the LR network ($L_{pa_{LR}}$) and the HR network ($L_{pa_{HR}}$):

\begin{equation}
    L_{pa}= L_{pa_{LR}}+L_{pa_{HR}}.
\end{equation}

\subsection{$L_2$ Reconstruction Loss}

$L_2$ reconstruction loss measures the reconstruction error in terms of Euclidean distance between the synthesized image and the corresponding real image and is defined for the LR and the HR network as follows:

\begin{equation}
    L_{2_{LR}}=\left\lVert G_{LR}(z|x^i_{LR})-y^i\right\rVert^2_2
\end{equation}
\begin{equation}
    L_{2_{HR}}=\left\lVert G_{HR}(z|x^j_{HR})-y^j\right\rVert^2_2.
\end{equation}
The total $L_2$ reconstruction loss function is given by:
\begin{equation}
    L_{2}=L_{2_{LR}}+L_{2_{HR}}.
\end{equation}

\subsection{Overall Objective Function}
The overall objective function for learning the network parameters in the proposed method is given as the sum of all the above defined loss functions:
\vspace{-0.25cm}
\begin{equation}
    \begin{split}
       L_{tot}&=L_{cpl}+ \lambda_1 L_a+ \lambda_2 L_{GAN} \\ & + \lambda_3 L_{P_{LR}}+\lambda_4 L_{pa}+ \lambda_5 L_2,
    \end{split}\label{eq:20}
\end{equation}
where $L_{cpl}$ is the coupling loss function, $L_a$ is the total attribute prediction loss function, $L_{GAN}$ is the total generative adversarial loss function, $L_{P_{LR}}$ is the perceptual loss for the LR network, $L_{pa}$ is the total perceptual attribute loss function and $L_2$ is the total reconstruction error. $\lambda_1, \lambda_2, \lambda_3, \lambda_4, \lambda_5$ are the adjustable hyper-parameters to weigh the different loss terms. 

\section {Experiments and Results}
In this section, we demonstrate the effectiveness of the
proposed approach by conducting various experiments on
four datasets: Labeled Faces in the Wild-a (LFWA) \cite{LFWa}, CelebFaces Attributes Dataset (CelebA) \cite{celebA}, Surveillance Camera Face (SCFace) \cite{Grgic2009SCfaceS}, and UnConstrained College Students (UCCS) Dataset \cite{UCCS}. We have compared our proposed method with six state-of-the-art methods on different datasets: VLRR \cite{wang2016studying}, DCA \cite{haghighat2017low}, LRFRW \cite{Li_2019_LRFRW}, D-align \cite{D-align_2019_TIFS}, SHSR \cite{SHSR} and SKD \cite{SKD}. In addition, we conduct an ablation study to demonstrate the effectiveness of each loss function of our network.


\subsection{Datasets}

CelebA consists of 202,599 images with training, validation and test splits of approximately 162,000, 20,000 and 20,000 images, respectively. The total dataset corresponds to about 10,000 identities (20 images per identity) with no identity overlap. Images are annotated with 40 facial attributes such as, ``wavy hair", ``chubby", ``bald", ``male", etc. However, we only use 12 attributes (shown in Table \ref{table:attr_pred}) for our proposed method. We use the pre-cropped version of the dataset, where the face images aligned using the hand-labeled key points. The image size in regular HR resolution is equal to $178 \times 212$. We downsample the images to low-resolutions of  $88 \times 108$, $68 \times 84$, $48 \times 58$. 

LFWA has a total of 13,232 images of 5,749 identities with pre-defined train and test splits dividing the entire dataset into approximately two equal partitions. Each image is annotated with the same 40 attributes used in CelebA dataset. The images are normalized to $224 \times 224$ for HR image and downsampled to $96\times 96$, $64\times 64$, $32 \times 32$. 

The SCface Dataset consists of 130 subjects, each
having one HR frontal face image and multiple HR images, captured from three distances (4.2m, 2.6m and 1.0m, respectively) using different quality surveillance cameras. For fair comparison with previous methods \cite{haghighat2017low}, 50 subjects are randomly selected for training and the rest 80 subjects for testing. As in \cite{haghighat2017low}, we fix the HR image at $128 \times 128$ and downsample to $64 \times 64$, $32 \times 32 $ and $16 \times 16 $ for LR images.

The UCCS dataset is a very challenging dataset taken under unconstrained conditions. Following the experimental setting in \cite{wang2016studying}, we perform evaluations on a 180-subject subset, where each subject has 25 or more images. We get a total of 5,220 images and  and use 4,200 images for training, and the remaining 1,020 images for testing. For fair comparison, we normalize the cropped face regions to $80 \times 80$ as HR, and downsample them for LR images of $16 \times 16 $. For the datasets SCFace and UCCS, which are not annotated with attributes, we use the state-of-the-art mixed-objective optimization network (MOON) \cite{Rudd2016MOON} for generating the ground truth attributes.

\subsection{Training Details}

As mentioned, we have implemented the U-Net network as generators and patch based discriminators for the LR and HR networks. The entire architecture has been been implemented in Pytorch. For convergence, all the hyper-parameters are set to 1 except  $\lambda_3$, and $\lambda_4$, which are set to 0.5. We have used a batch size of 6 for Adam optimizer \cite{Kingma2015AdamAM} with first-order momentum of 0.5, and learning at a rate of 0.0004. We have used ReLU activation for the generator and Leaky ReLU with a slope of 0.25 for the discriminator. For fine-tuning the attribute predictor network VGG-Face for attribute perceptual loss, we have chosen 12 attributes (shown in Table \ref{table:attr_pred} from the LFWA dataset).  

The complex loss in (\ref{eq:20}) makes it difficult to train the whole network directly as the gradient diffusion caused by different tasks will lead to slow network convergence. To address this issue, we have employed a stage-wise learning strategy, where the information in the training data is presented to the network gradually. Specifically, we first optimize each task greedily by not updating the other task simultaneously. After the ‘initialization’ for each task, we fine-tune the whole network all together by optimizing all the tasks jointly. 

For training, we require genuine and impostor pairs. The genuine/impostor pairs are constructed using LR and HR images of same/different subject. We balance the training set by using same number of genuine and impostor pairs.

\begin{figure*}[h]
\centering     

\subfigure[LFWA ]{\label{fig:a}\includegraphics[width=4.5cm,height=4.1cm]{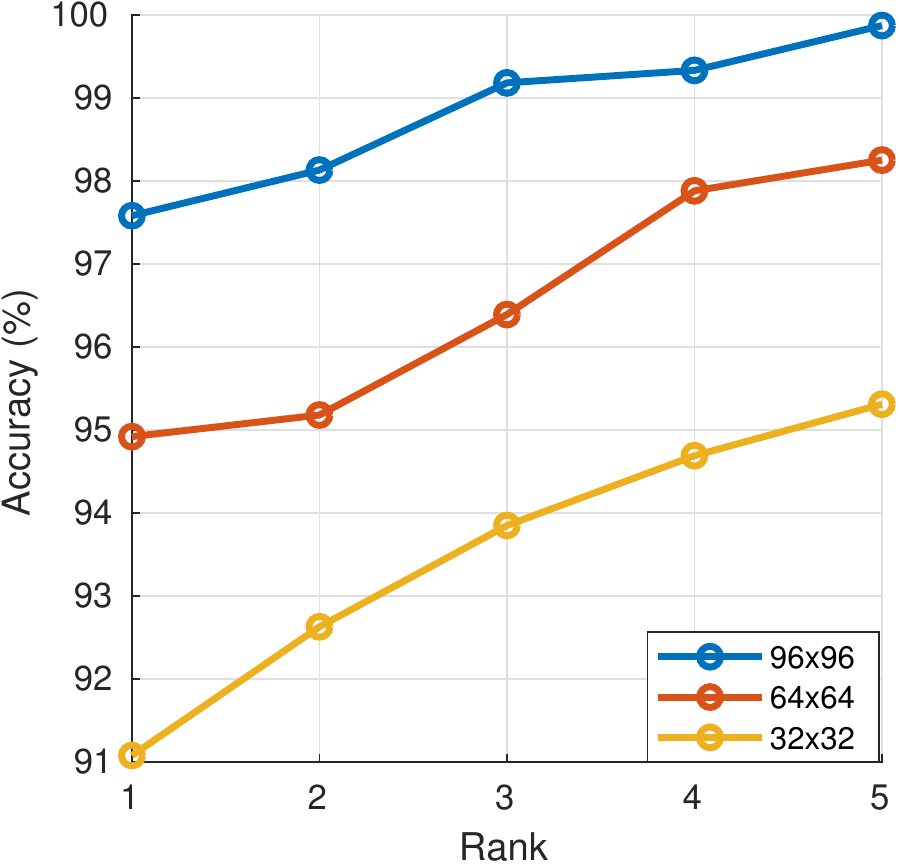}}
\subfigure[CelebA]{\label{fig:b}\includegraphics[width=4.5cm,height=4.1cm]{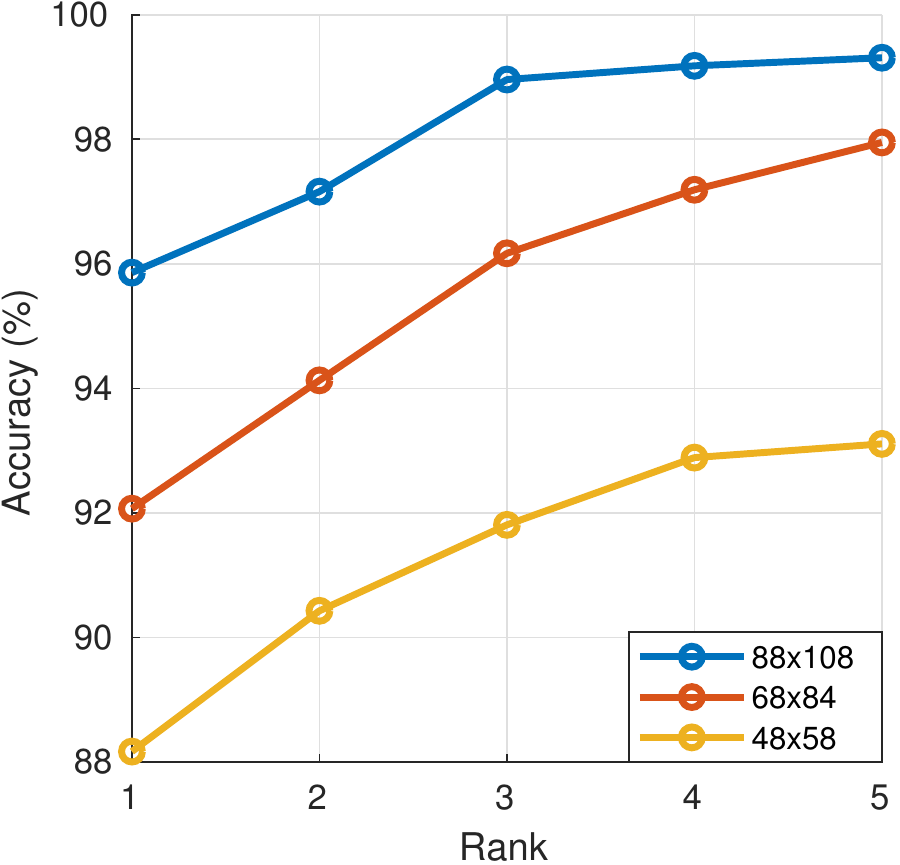}}
\subfigure[SCFace]{\label{fig:c}\includegraphics[width=4.3cm,height=4.1cm]{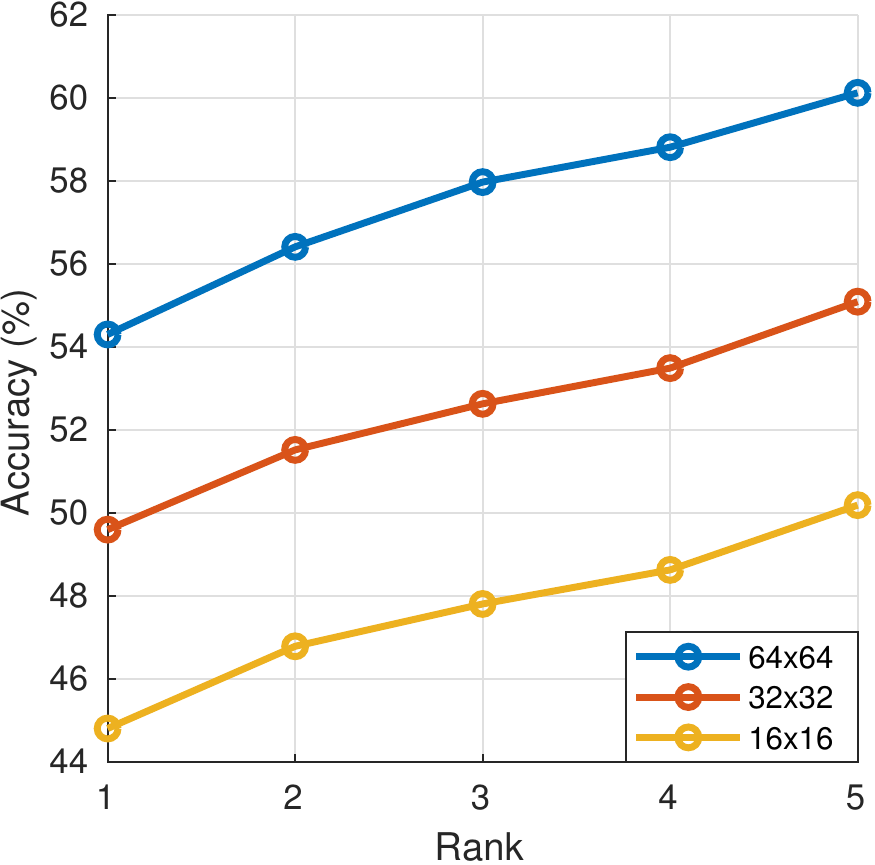}}

\vspace{-0.25cm}
\caption{CMC curves for rank-n recognition accuracy for different low-resolution images for different datasets.}
\label{fig:cmc}
\vspace{-0.30cm}
\end{figure*}


 


\begin{figure}[h]
\centering
\includegraphics[width=5.65cm]{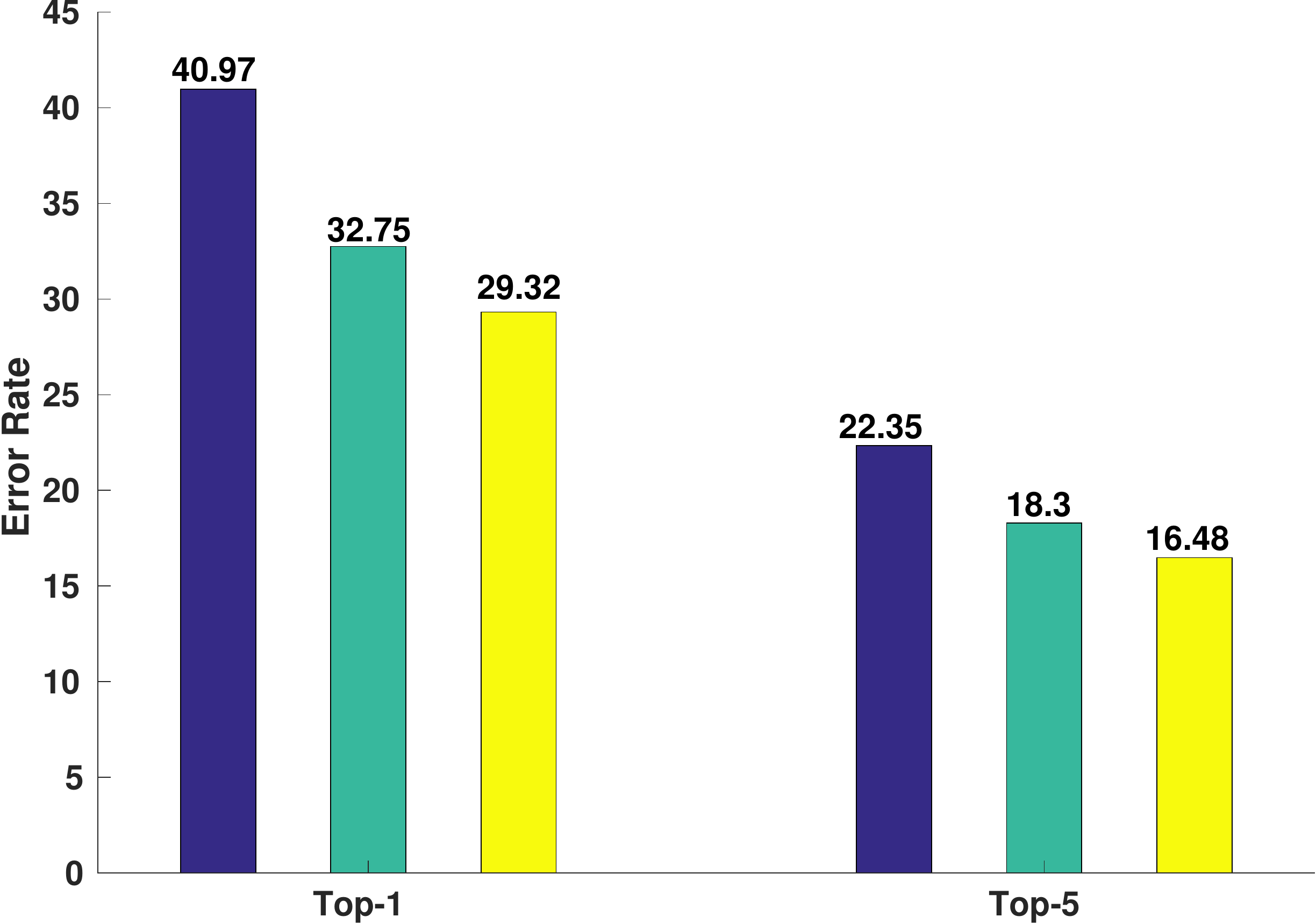}
\caption{Top-1 and Top-5 Error rate comparison for VLLR (blue), SKD (Green), and our method (Yellow) using the UCCS dataset.}\label{fig:bar_UCCS}

\end{figure}
\begin{figure}[h]
\centering
\includegraphics[width=5.65cm]{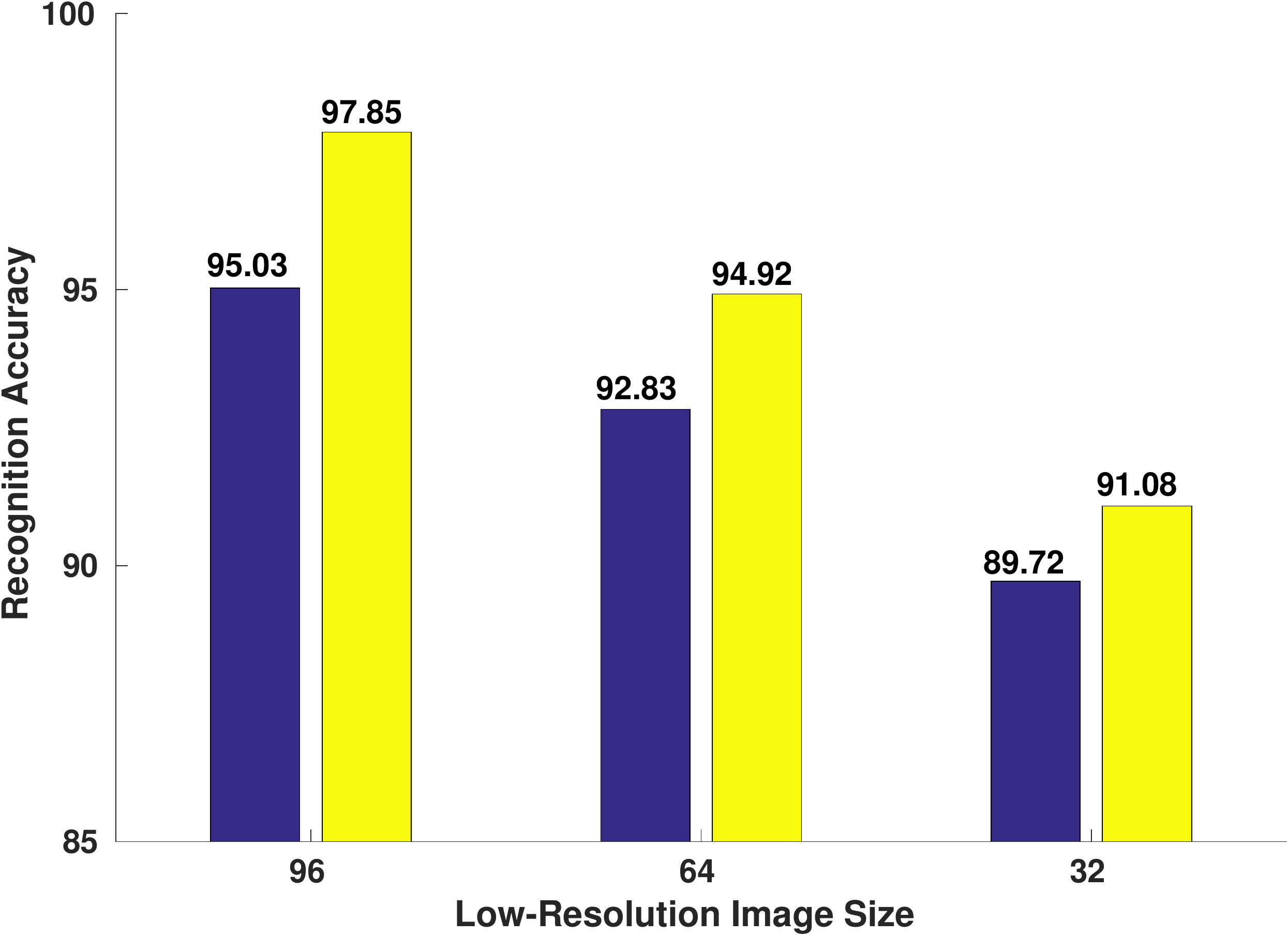}
\caption{Recognition accuracy ($\%$) comparison for SKD (Blue), and our method (Yellow) using the LFWA dataset for different size of LR image.}\label{fig:bar_LFWA}
\vspace{-0.35cm}
\end{figure}
\subsection{Testing of the Proposed Method}

The main objective of our proposed method is to match a test LR image against a gallery of HR images using the corresponding feature set from the common latent embedding subspace. During testing, a given probe LR image $x^p_{LR}$ is passed through the LR network, and $z_1(x^p_{LR})$ is generated from the common embedding subspace. Similarly, the HR images from the gallery are passed through the HR network and $z_2(x^j_{HR})$  is measured for each image $x^j_{HR}$. Eventually, the face recognition is performed by calculating the minimum Euclidean distance between $z_1(x^p_{LR})$ and $z_2(x^j_{HR})$ for all the gallery HR images:
\vspace{-0.22cm}
\begin{equation}
    \hat{j}=\operatorname*{arg\,min}_j \left\lVert z_1(x^p_{LR})-z_2(x^j_{HR}) \right\rVert_2^2.  
\end{equation}   
Therefore, $x^{\hat{j}}_{HR}$ is the matching HR image from the gallery for the given probe LR image $x^p_{LR}$. The ratio of the number of correctly classified probes to the total number of probes is computed as the identification rate.

Additionally, the LR network can also be used for facial attribute prediction of a given LR probe image by passing the feature set $z_1(x^p_{LR})$ through the attribute predictor of the LR network. The predicted facial attribute can be used to narrow down the search for identification in a large gallery of HR images.


\subsection{Performance Evaluation}

We have evaluated the proposed method and compared with other state-of-the-art methods on four different datasets using different low-resolution images. Fig. \ref{fig:cmc} provides the recognition accuracy of our proposed method from rank-1 to rank-5 for different resolution images using the LFWA, CelebA and SCFace dataset. We can clearly see that the proposed method gives very good performance for the LFWA and CelebA dataset. However, the SCFace  has more challenging face variations than the LFWA and CelebA  and the SCFace images have been taken in a typical commercial surveillance environment, which leads to lower recognition performance for the SCFace dataset when compared to LFWA and CelebA as seen in Fig. \ref{fig:cmc}.

\begin{table}[t]
\centering
\captionsetup{width=.95\linewidth}
\caption{Rank-1 recognition accuracy ($\%$) on SCFace.}
\scalebox{0.75}{\begin{tabular}{c c c c c}
 \hline
Model & Dist-1& Dist-2&Dist-3& Average \\ \hline
SHSR&14.70&15.70&19.10&16.50 \\ \hline
DCA&12.19&18.44&25.53&18.72 \\ \hline
LRFRW&20.40&20.80&31.71&18.72 \\ \hline
D-Align&34.37&39.38&49.37&24.30 \\ \hline
SKD&43.50&48.00&53.50&48.33 \\ \hline
Our Method& $44.81 \pm 0.36$& $49.60 \pm 0.41 $ &$54.30 \pm 0.23$&$49.57 \pm 0.39$ \\ \hline\end{tabular}}
\label{table:table_scface}
\end{table}

\begin{figure}
\centering
\includegraphics[width=6.5cm]{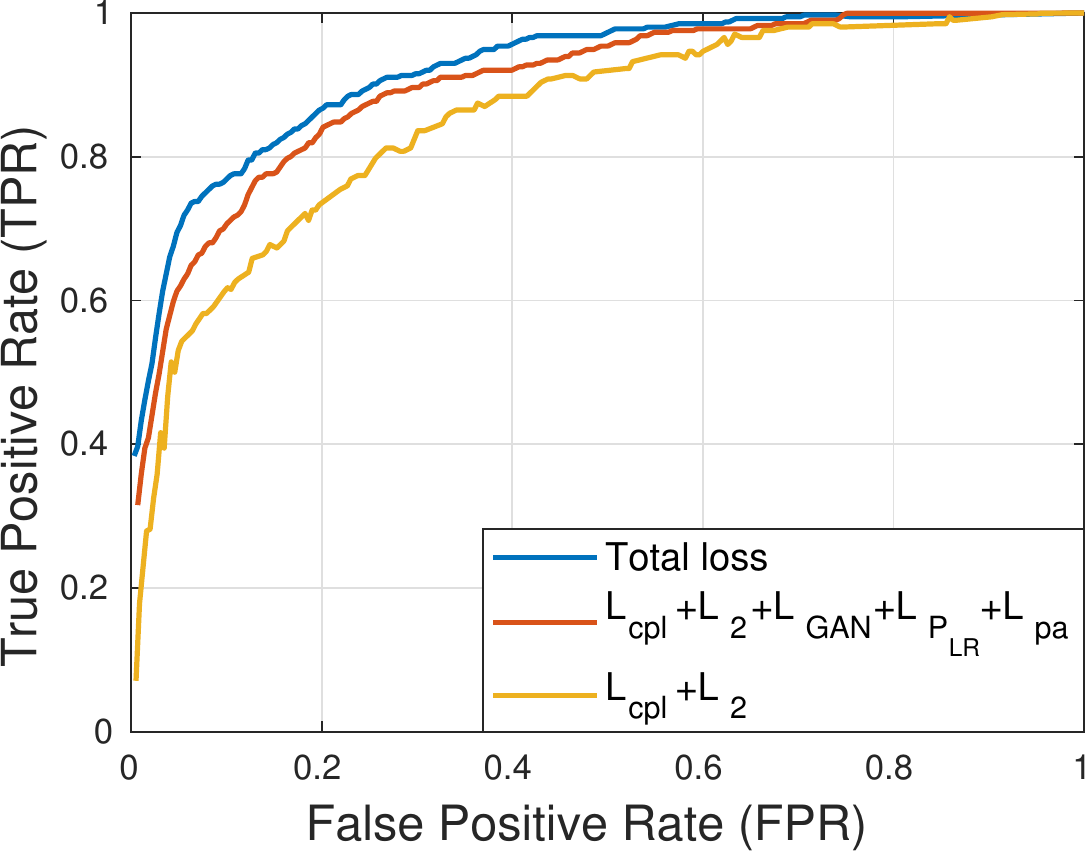}
\caption{ROC curves corresponding to the ablation study.}\label{fig:ROC}
\vspace{-0.35cm}
\end{figure}

\begin{table*}[t]
\centering
\captionsetup{width=.95\linewidth}
\caption{Attribute prediction accuracy ($\%$) comparison using CelebA dataset.}
\scalebox{0.67}{\begin{tabular}{c|c|c| c| c| c| c| c| c| c| c| c| c }
 \hline
  & Double Chin & Chubby & Eye Glasses & Male & Pale Skin & Moustache & Mouth Slightly Open & Young & Smiling & Goatee & Bald & Blond Hair  \\ \hline
HR Input (Net A)&90.8&90.1&96.5&97.5 &89.2&92.8&90.1&85.3&91.3&93.8&97.4&92.7 \\ 
LR Input (Net A)&51.3&50.1&56.5& 55.3&46.4&53.1&51.6& 49.2&55.4&56.9&62.8&59.7 \\ 
LR Input (fine-tuned Net A)&71.6&69.3&74.6&72.9&68.6&71.4&70.3&69.5&75.8&76.0&88.5&81.4 \\ 
LR Input (Proposed Method) &83.2&82.6&89.3&91&83.3&88.1&86.6&78.9&87.2&88.9&93.6&88.4 \\ \hline\hline
\end{tabular}}
\label{table:attr_pred}
\vspace{-0.3cm}
\end{table*}

Table \ref{table:table_scface} tabulates the comparison of Rank-1 recognition rates with different state-of-the-art methods for the SCFace dataset using different distances from the camera, which corresponds to different resolution images. We can observe that our proposed method outperforms the state-of-the-art embedded method of cross-resolution face recognition model SKD \cite{SKD} by $1.31\%$,$1.60\%$,$0.80\%$, and $1.22\%$ on Dist-1, Dist-2, Dist-3, and average, respectively. We can also observe that our model even outperforms the state-of-the-art hallucination model SHSR \cite{SHSR} by approximately $33\%$ an average for all the three distances. 
We have also compared proposed method with VLRR \cite{wang2016studying}, and SKD \cite{SKD} using the UCCS dataset. Top-1 and Top-5 error rate comparison has been shown in Fig. \ref{fig:bar_UCCS}. UCCS is also a very challenging dataset, where the faces have been captured in completely unconstrained conditions. Due to this reason, the error rates for this dataset are very high. However, our proposed method performs better than the other two compared method by giving a lower error rate of at least $3.4\%$ and $1.8\%$ for Top-1 and Top-5 recognition. We can also notice from Fig. \ref{fig:bar_LFWA}, that our proposed method outperforms SKD even for the LFWA dataset for different resolutions.

From performance evaluation, we observe that our proposed coupled framework with the contrastive loss function and leveraging facial attributes to transform different domains (LR and HR) into a common discriminative embedding subspace is superior than the other embedding techniques such as SKD and D-Align. It also shows the efficacy of exploiting multiple loss functions for cross-resolution face recognition. The relative importance of the loss functions has been covered in detail in ablation study (Sec. \ref{subsec:ablation}).

\subsection{Attribute Prediction for Low-Resolution}
One of the advantages of the proposed method is that it can be used for attribute prediction for LR face images. To illustrate the efficacy of our proposed approach for attribute prediction of LR face images, we have performed attribute prediction for 4 different scenarios: 1) Attribute prediction for HR images with the VGG-Face based attribute predictor, which is represented as Net A in Sec. \ref{subsec:percloss}. 2) Attribute prediction for LR images using attribute predictor Net A. 3) In this scenario, we first fine-tune the attribute predictor A with annotated LR images and then use it for attribute prediction of LR test images. This will be called ``fine-tuned Net A" 4) In this final case, we test our attribute predictor from LR network for attribute prediction of the LR test images (see Fig. \ref{fig:enrol}). We have performed this experiment for the Celeb-A dataset using $68 \times 88$ as LR images.

The attribute prediction results for the above  4 scenarios for 12 attributes using the Celeb-A  have been tabulated in Table \ref{table:attr_pred}. We can notice from Table \ref{table:attr_pred} that our  approach shows the best performance in predicting the attributes of LR images for both datasets. Fine-tuning the Net A with LR images helps in improving its performance, however it does not perform as well as our method. Additionally, the performance of our LR network attribute predictor is comparable to the Net A performance for HR images.
\subsection{Ablation Study}\label{subsec:ablation}
The objective function defined in (\ref{eq:20}) contains multiple loss functions: coupling loss ($L_{cpl}$), attribute prediction loss ($L_a$), perceptual loss ($L_{P_{LR}}$, $L_{pa}$), $L_2$ reconstruction loss ($L_2$), and GAN loss ($L_{GAN}$). In this section, we study the relative importance of different loss functions and the benefit of using them in our proposed method. For this experiment, we use different variations of our proposed approach and perform the evaluation using the LFWA dataset ($64 \times 64$ LR images). The variations are: 1) cross-resolution face verification using the coupled framework with only coupling loss and $L_2$ reconstruction loss ($L_{cpl} + L_2$); 2) cross-resolution face verification using the coupled framework with coupling loss, $L_2$ reconstruction loss, GAN loss and perceptual loss ($L_{cpl} + L_2 + L_{GAN} + L_{P_{LR}} + L_{pa}$); 3) cross-resolution face verification using our framework with all the loss functions ($L_{cpl} + L_2 + L_{GAN} + L_{P_{LR}} + L_{pa} + L_{a}$).

We use the above three variations of our framework and plot the receiver operating characteristic (ROC) curve for the task of cross-resolution face verification using the features from the common embedding subspace. We can see from Fig. \ref{fig:ROC} that the generative adversarial loss and the perceptual loss (red curve) help in improving the cross-resolution verification performance, and adding the attribute prediction loss (blue curve) helps in more performance improvement. The reason for this improvement is that using facial attribute loss along with the contrastive loss leads to a more discriminative embedding subspace leading to a better face recognition performance. This also shows that multitask learning of attribute prediction and face recognition is useful and helps in cross-resolution face recognition task.

\section{Conclusion}
   We have proposed a novel framework which adopts a coupled GAN and exploits facial attributes for cross-resolution face recognition. The coupled GAN includes two sub-networks which project the low and high-resolution images into a common embedding subspace, where the goal of each sub-network is to maximize  the  pair-wise  correlation  between low and high-resolution images during the projection process. Moreover, we leverage facial attributes to further maximize the pair-wise correlation by  implicitly matching facial  attributes  of  the  low  and  high-resolution images during the training. We  comprehensively evaluated our model on four standard datasets and the results indicate that our model significantly outperforms other state-of-the-art models for cross resolution face recognition. Additionally, the enhancement obtained by different losses in the proposed method has been considered in an ablation study.
{\small
\bibliographystyle{ieee}
\bibliography{submission_example}
}

\end{document}